# Summary of Point Transformer with Federated Learning for Predicting Breast Cancer HER2 Status from Hematoxylin and Eosin-Stained Whole Slide Images

Kamorudeen A. Amuda
*Department of Computer and Information Sciences*
*Towson University*
Maryland, USA
Email: kamuda@towson.edu
ORCID: 0000-0001-5646-1867

Almustapha A. Wakili
*Department of Computer and Information Sciences*
*Towson University*
Maryland, USA
Email: awakili@towson.edu

*Abstract*—This study introduces a federated learning-based approach to predict HER2 status from hematoxylin and eosin (HE)-stained whole slide images (WSIs), reducing costs and speeding up treatment decisions. To address label imbalance and feature representation challenges in multisite datasets, a point transformer is proposed, incorporating dynamic label distribution, an auxiliary classifier, and farthest cosine sampling. Extensive experiments demonstrate state-of-the-art performance across four sites (2687 WSIs) and strong generalization to two unseen sites (229 WSIs).

Introduction

Hematoxylin and eosin (HE)-stained whole slide images (WSIs) are increasingly used beyond traditional tasks, applying deep learning to detect subtle molecular characteristics [1], [2]. In breast cancer, accurately predicting the status of HER2 is critical to guide treatment decisions [3]. Typically, HER2 status determination is based on immunohistochemistry (IHC) or in situ hybridization (ISH) [4], but deep learning enables prediction directly from HE-stained WSIs, reducing the need for these costly techniques.

Federated learning (FL) [5] has shown promise in WSI analysis by enabling collaboration between multiple sites without the need to transfer large datasets, minimizing concerns about data privacy. However, challenges such as label imbalance and variations in specimen preparation across sites can impact model performance. While existing methods address these issues in natural scenes, applying them to WSI datasets presents additional challenges compared to centralized learning.

The proposed PointTransformerDDA+ approach addresses these challenges by capturing both local context and longrange dependencies using a point transformer block enriched with positional information. The proposed Farthest Cosine Sampling (FCS) method identifies the most distinctive features to capture long-range dependencies. Additionally, the proposed Dynamic Distribution Adjustment (DDA) method tackles label imbalance across sites, improving model initialization and generalization.

Key contributions of the proposed method include:
- Using a point transformer to effectively capture both local context and long-range dependencies in WSI analysis.
- Implement FCS to capture long-range dependencies and aggregate distinctive features.
- Applying DDA to manage class imbalance across multiple sites, enhancing generalization.
- Demonstrating state-of-the-art performance for the prediction of HER2 in four sites (2687 WSIs) and generalizing to two unseen sites (229 WSIs).

Related Work

Recent works on WSI classification, including HER2 status prediction, have primarily used MIL-based or graph-based methods. MIL-based approaches often utilize attention mechanisms or transformers to capture long-range dependencies but may neglect local spatial information. DSMIL combines features from different scales, while TransMIL models spatial relationships using transformers with position encoding not based on Euclidean space. Graph-based models, like PatchGCN and SlideGraph+, capture local information via graph structures but may struggle with long-term dependencies. Point neural networks, which capture both local context and longrange dependencies, have seen limited application in WSI classification [1], [2].

Federated Learning in WSI Analysis

Federated learning [5], [6] enables multi-site WSI model training but faces challenges like skewed label distribution. HistFL and TNBC-FL apply federated learning for cancer predictions, while approaches like FedProx and FedMGDA+ improve robustness on non-i.i.d. data. Despite progress, federated learning still underperforms compared to centralized learning for WSIs, needing further enhancement [6], [7].

Methodology

This section introduces the problem of HER2 status prediction using a point transformer with federated learning.

We describe the main components of the framework, including point feature extraction, the point transformer block, the point abstraction block, and federated learning with dynamic distribution adjustment. Figure 1 provides an overview of the proposed framework, highlighting the repeated blocks and key components such as the feature pyramid network (FPN), global average pooling (GAP), and multilayer perceptron (MLP).

Point Feature Extraction

To extract the features of points from a WSI, the CLAM method [1] is used to pre-process and patch the WSIs, where each patch is treated as a point. The coordinates of each patch are represented as ($px,py,1$), where the z-coordinate is fixed. These patches are then inputted into a nuclei segmentation network, pre-trained using a Swin-Transformer with a fourlevel feature pyramid network (FPN). The outputs from the four FPN layers are averaged and concatenated with the patch coordinates, resulting in a patch-level feature $x_n \in R^{3+d}$, where $d = 256$. For each WSI with label $y$, the point set $X = \{x_1, x_2, ..., x_{|X|}\}$ is formed. To optimize memory and computation, 1024 patches are randomly selected from each WSI.

Point Transformer Block

In our approach, we use the original point transformer block [8] to capture and aggregate local context information for each point via an attention mechanism. For each point $x_i$ with position $p_i$, we consider its $k$-nearest neighbors (where $k = 16$) as a subset $X^{(i)} \subset X$. The attention score $\alpha(x_i, x_j)$ between $x_i$ and its neighboring points $x_j \in X^{(i)}$ is calculated as:

$$\alpha(x_i, x_j) = W_q(W_i x_i - W_j x_j) + PE(p_i, p_j)$$

where $PE(p_i, p_j)$ is the relative position encoding:

$$PE(p_i, p_j) = MLP(p_i - p_j)$$

After computing the attention scores, the localized features around $x_i$ are aggregated using a softmax function, and the final output $y_i$ is obtained by applying a residual connection:

$$z_i = \sum_{x_j \in X^{(i)}} S(\alpha(x_i, x_j)) \cdot (W_v x_j + PE(p_i, p_j))$$

$$y_i = x_i + W_z z_i$$

This approach effectively aggregates local point features using attention and relative positional encoding.

Point Abstraction Block

To reduce the cardinality of a point set and capture longrange dependencies, a novel strategy called farthest cosine sampling (FCS) is proposed, replacing farthest point sampling (FPS). Unlike FPS, which samples on the basis of position, FCS operates in the feature space, using the cosine distance as the metric between points. This method prevents missing important positive patches, especially in cases where WSIs contain few clustered positive patches. The cosine distance is defined as:

$$\text{Dist}(x_i, x_j) = 1 - \frac{x_i \cdot x_j}{\max(\|x_i\|_2 \cdot \|x_j\|_2, 1e-8)}$$

FCS iteratively selects the $M/4$ farthest points, ensuring that important patches are retained. After FCS, a k-nearest neighbor grouping is applied, followed by a max pooling of the neighborhood features for each sampled point:

$$y_i = \text{MaxPooling}_{x_j \in X_2(i)}(MLP(x_j))$$

This approach improves HER2 status prediction by covering key points and capturing long-range dependencies in the feature space.

Point Classifier Block After performing four attention and abstraction operations, four abstract points are obtained, represented as $F_g \in R^{4 \times 512}$. By averaging these grouped features, the final WSI-level feature $F_h \in R^{64}$ is derived:

$$F_h = MLP(GAP(F_g)) = f_h(X)$$

The MLP has two layers, each with a linear transformation and ReLU activation. A linear layer with a softmax function, denoted as $f_c$, outputs the final HER2 status probabilities $p$. The loss is computed using the cross-entropy (CE) loss function:

$$\ell(p, y) = -\frac{1}{|P|} \sum_{i=1}^{|P|} \sum_{c=0}^{1} y_{ic} \log(p_{ic})$$

This equation calculates the prediction error based on the HER2 status.

Federated Learning with Dynamic Distribution Adjustment

To address the non-i.i.d. label distribution issue in WSIs from different sites, a dynamic distribution adjustment strategy is introduced. This strategy subsamples the majority of HER2WSIs to balance the label distribution in the early stages of federated learning. Over time, the balanced distribution gradually adjusts to match the real distribution. Specifically, all HER2+ WSIs are kept, and a Bernoulli mask $M(k)$ is applied to the HER2- WSIs, where the subsampling probability $b_k$ evolves with training iterations. The loss function is formulated as:

$$L_{cls} = M \cdot \ell(f_c(F_h), y)$$

To prevent potential information loss from subsampling, an auxiliary classifier is added, which incorporates the real label distribution of each site.

$$L_{aux}(p, y) = -\frac{1}{|P_i|} \left[ \sum_i \log(p_i^0) + \sum_i \log(p_i^1) \right]$$

The total loss is then computed as

$L_{total} = L_{cls} + L_{aux}$

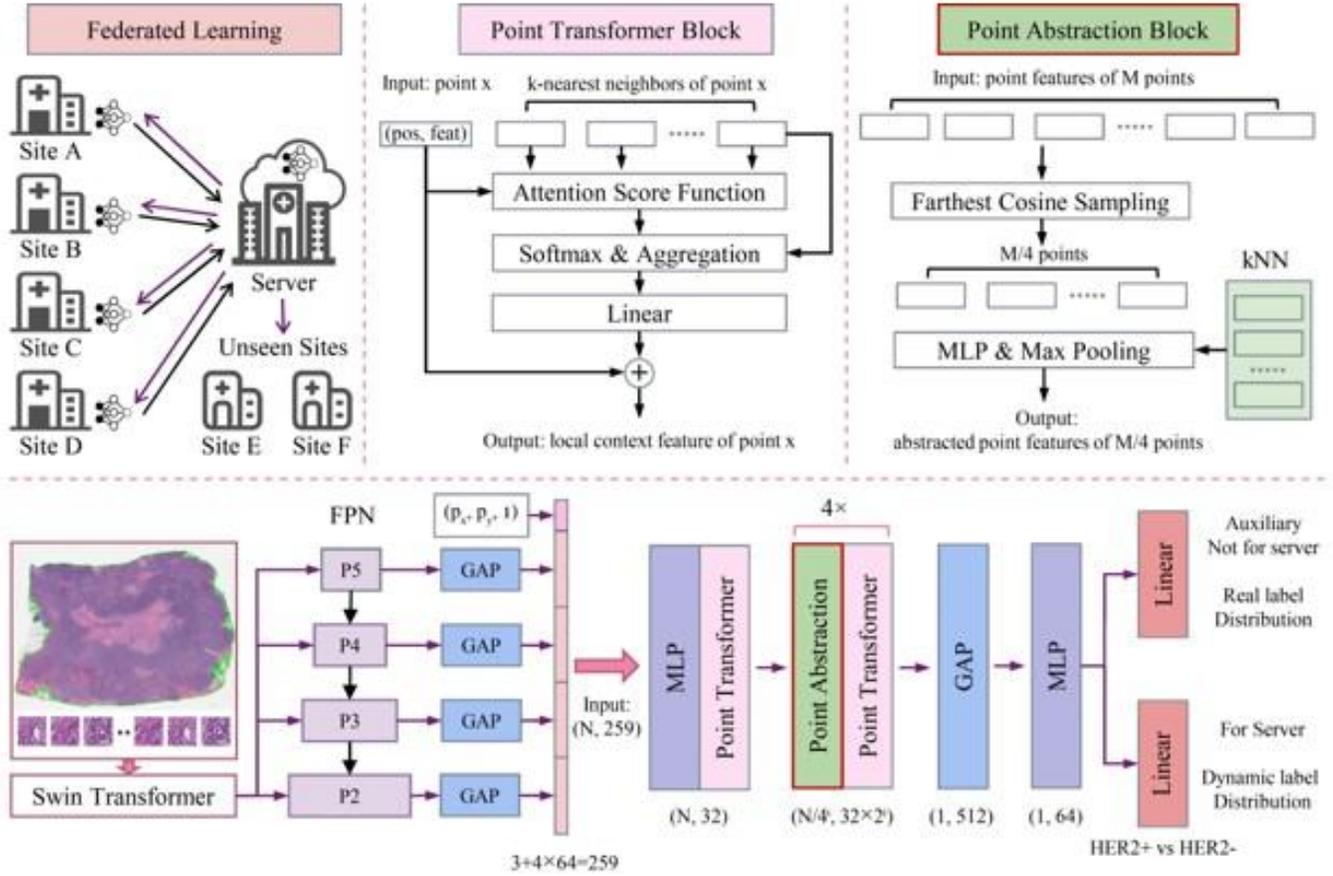

Fig. 1: Point transformer for predicting HER2 status from whole slide images in a federated learning framework

The auxiliary classifier improves feature representation but is not involved in weight synchronization during federated learning. This approach ensures a well-initialized model and reduces the impact of imbalanced data.

Experiments

Dataset and Experimental Settings

The point transformer model for HER2 status prediction was tested on 2,916 WSIs from six sites. Four sites were used for federated learning, while two served as unseen data for external testing. Data from the federated sites were split into training (60%), validation (10%), and test (30%) sets. The evaluation was based on the mean AUC from the test set, using five repeated splits for model selection. Three model variants were tested: PointTransformer+ (with FCS), PointTransformerDDA (with DDA), and PointTransformerDDA+ (combining FCS and DDA).

Comparison with WSI Classification Methods

The evaluation includes models such as PointNet++, MILbased models (CLAM-SB, DSMIL, TransMIL, HistFL), and graph-based models (GraphSAGE, Patch-GCN, SlideGraph+) for comparison, all using federated learning. Table 1 shows that point-based models perform competitively by integrating local neighborhood features and capturing long-range dependencies, similar to both MIL and graph-based methods. The introduction of FCS and DDA strategies allows the point transformer to outperform other models, with PointTransformerDDA+ achieving state-of-the-art AUC on the test set and federated sites. TransMIL also delivers strong performance, highlighting the importance of position encoding in HER2 status prediction.

Comparison with Federated Learning Methods

The point transformer model's performance with different federated learning methods is evaluated, with PointTransformerDDA+ achieving the highest total AUC, closely matching centralized training. FCS helps capture discriminative features, while DDA addresses non-i.i.d. issues, both improving federated learning performance. Although GroupNorm [9] performs well at Site D, it is sensitive to the data and requires careful tuning. Despite not being designed for unseen scenarios, the model still delivers strong results AUC ¿ 0.79 for two unseen sites.

Conclusion

This study presents a novel approach for Whole Slide Image (WSI) analysis by treating a WSI as a point cloud with position data, rather than using traditional MIL or graphbased methods.

A farthest cosine sampling method is introduced to capture long-range dependencies and aggregate key point features. To address label imbalance in real-world WSIs under federated learning, a dynamic distribution adjustment is proposed. Experiments show the effectiveness of these components, achieving strong performance on unseen sites and IHC score 2+ subsets.

TABLE I: Comparison of our model with other points, multi-instances, and graph-based models.

| Experiment | Method | Average | Site A | Site B | Site C | Site D |
|---|---|---|---|---|---|---|
| | PointTransformerDDA+ | 0.816 ± 0.019 | 0.766 ± 0.025 | 0.866 ± 0.021 | 0.837 ± 0.036 | 0.760 ± 0.046 |
| Ours | PointTransformerDDA | 0.793 ± 0.013 | 0.730 ± 0.029 | 0.855 ± 0.013 | 0.804 ± 0.024 | 0.758 ± 0.022 |
| | PointTransformer++ | 0.806 ± 0.015 | 0.752 ± 0.018 | 0.844 ± 0.020 | 0.823 ± 0.024 | 0.757 ± 0.019 |
| Point-based | PointTransformer [1] | 0.771 ± 0.037 | 0.717 ± 0.037 | 0.834 ± 0.026 | 0.776 ± 0.037 | 0.721 ± 0.035 |
| | PointNet++ [2] | 0.763 ± 0.043 | 0.696 ± 0.048 | 0.830 ± 0.032 | 0.782 ± 0.045 | 0.730 ± 0.042 |
| MIL-based | CLAM-SB [3] | 0.767 ± 0.032 | 0.712 ± 0.044 | 0.793 ± 0.037 | 0.766 ± 0.072 | 0.748 ± 0.022 |
| | DSMIL [4] | 0.693 ± 0.021 | 0.648 ± 0.052 | 0.751 ± 0.031 | 0.675 ± 0.080 | 0.675 ± 0.011 |
| | TransMIL [5] | 0.739 ± 0.019 | 0.739 ± 0.035 | 0.824 ± 0.021 | 0.805 ± 0.036 | 0.616 ± 0.035 |
| | HistoFL [6] | 0.757 ± 0.045 | 0.729 ± 0.062 | 0.776 ± 0.050 | 0.812 ± 0.053 | 0.713 ± 0.041 |
| Graph-based | GraphSAGE [7] | 0.711 ± 0.047 | 0.692 ± 0.050 | 0.738 ± 0.022 | 0.692 ± 0.048 | 0.685 ± 0.047 |
| | Patch-GCN [8] | 0.750 ± 0.032 | 0.716 ± 0.039 | 0.768 ± 0.062 | 0.766 ± 0.030 | 0.728 ± 0.047 |
| | SlideGraph+ [9] | 0.783 ± 0.019 | 0.736 ± 0.029 | 0.825 ± 0.020 | 0.804 ± 0.027 | 0.785 ± 0.013 |